\documentclass{article} 
\usepackage{iclr2016_conference,times}
\usepackage{hyperref}
\usepackage{url}
\usepackage{amsmath}
\usepackage{amsthm}
\usepackage{amsfonts}
\usepackage{graphicx}
\usepackage{subcaption}
\usepackage{algorithm2e}
\usepackage{multirow}
\usepackage{pbox}

\newtheorem{theorem}{Theorem}

\DeclareMathOperator{\argmin}{argmin}

\title{Task Loss Estimation for Sequence Prediction}

\author{Dzmitry Bahdanau, Dmitriy Serdyuk, Phil\'emon Brakel,
   Nan Rosemary Ke \\
   Universit\'e de Montr\'eal\\
\And
Jan Chorowski \\
University of Wroc\l{}aw \\
\And
Aaron Courville, Yoshua Bengio\thanks{Yoshua Bengio is a CIFAR Senior Fellow}
\\
Universit\'e de Montr\'eal
}

\begin{document}

\maketitle

\begin{abstract}
Often, the performance on a supervised machine learning task is evaluated with a
\emph{task loss} function that cannot be optimized directly. Examples of such loss functions
include the classification error, the edit distance and the BLEU
score. A common workaround for this
problem is to instead optimize a \emph{surrogate loss}
function, such as for instance cross-entropy or hinge loss. In
order for this remedy to be effective, it is
important to ensure that minimization of the surrogate loss
results in minimization of the task loss, a condition
that we call \emph{consistency with the task loss}.
In this work, we propose another method for deriving
differentiable surrogate losses that provably meet this requirement.
We focus on the broad class
of models that define a score for every input-output pair.
Our idea is that this score can be interpreted as an
estimate of the task loss, and that the estimation error may 
be used as a consistent surrogate loss. A distinct feature of
such an approach is that it defines the desirable value of the
score for every input-output pair. We use this property 
to design specialized surrogate losses for  Encoder-Decoder
models often used for sequence prediction tasks.
In our experiment, we benchmark on the task of speech
recognition. Using a new surrogate loss instead of
cross-entropy to train an Encoder-Decoder speech recognizer
brings a significant ~13\% relative improvement in 
terms of Character Error Rate (CER) in the case when no 
extra corpora are used for language modeling.
\end{abstract}

\section{Introduction}

There has been an increase of interest in learning systems that can
solve tasks in an ``end-to-end'' fashion.
An early example of such a system is a highly successful convolutional network handwriting
recognition pipeline \citep{lecun1998gradient}. More recent examples are deep
convolutional networks designed for image recognition
\citep{krizhevsky2012imagenet}, neural translation systems
\citep{sutskever2014sequence,bahdanau2015neural}, and speech
recognizers
\citep{graves2014towards,hannun2014deep,chorowski2015attention,bahdanau2015end-to-end}.
Parts of end-to-end systems, such as image features
extracted by convolutional networks, often successfully replace
hand-designed ones \citep{yosinski2014how}. This
demonstrates how useful it can be that all parts of a system
are learned to solve the relevant task.

In practice however, it often happens that the relevant
\emph{task loss} function, such as error rate in
classification, word error rate in speech recognition, or BLEU
score in machine translation, is only used for model
evaluation, while a different \emph{surrogate loss} is used
to train the model.
There are several reasons for the evaluation loss -- training loss
discrepancy: the evaluation criterion may be non-differentiable, it
can be non-convex or otherwise inconvenient to optimize, or one
may want to emphasize certain problem-agnostic model properties,
such as a class separation margin \citep{vapnik1998statistical}. For
instance, classification models are often evaluated based on their
error rates, which corresponds to a 0-1 task loss. However, people often minimize
surrogate losses like the cross-entropy \citep{bishop2006prml} or the hinge loss
\citep{vapnik1998statistical} instead. For classification, these surrogate
losses are well-motivated and their minimization tends to lead to a low
error rate. It is not clear, however, that the same methods
should be preferred for structured output problems, in which
typically there is a gradation in the quality of answers.



In this work, we revisit the problem of choosing an
appropriate surrogate loss for training.  We focus on the
broad class of models that define a score for every
input-output pair and make predictions by looking for the
output with the lowest score. Our main idea is that if the
scores defined by the model are approximately equal to the
task loss, then the task loss of the model's prediction
should be low.  We hence propose to define the surrogate
loss as the estimation error of a score function that is
trained to mimic the task loss, a method we will refer to as
\emph{task loss estimation}.  We prove that minimization of
such a surrogate loss leads to the minimization of the
targeted task loss as well, a property that we call
\textit{consistency} with the task loss. 
The main distinct feature of our new approach is that it
prescribes a target value for the score of every input-output
pair. This target value does not depend on the score of
other outputs, which is the key property of the proposed
method and the key difference from other approaches 
to define consistent surrogate losses, such as the generalized
hinge loss used in Structured Support Vector Machines
\citep{tsochantaridis2005large}. 

Furthermore, we apply the task loss estimation principle to
derive new surrogate losses for sequence prediction models
of the Encoder-Decoder family. The Decoder, typically a
recurrent network, produces the score for an input-output
pair by summing terms associated with every element
of the sequence. The fact that the target for the score is
fixed in our approach allows us to define targets for each
of the terms separately. By doing so we strive to achieve two
goals: to facilitate faster training and to ensure that the
greedy search and the beam search used to obtain predictions
from an Encoder-Decoder work reasonably well. To validate 
our ideas we carry out an experiment on a speech recognition
task. We show that when no external language model is used
using a new surrogate loss indeed results in a relative 13\%
improvement of the CER compared to cross-entropy training
for an Encoder-Decoder speech recognizer.

\section{Task Loss Estimation for Supervised Learning}\label{sec:lossdefs}

\paragraph{Basic Definitions}
Consider the broad class of supervised learning problems in
which the trained learner is only allowed to
deterministically produce a single answer $\hat{y} \in
\mathcal{Y}$ at run-time, when given an input $x \in
\mathcal{X}$. After training, the learner's performance
is evaluated in terms of the task loss $L(x, \hat{y})$ that it suffered
from outputting $\hat{y}$ for $x$. We assume that the task loss is non-negative
and that there
exists a unique ground truth answer $y=g(x)$ such that $L(x,
g(x))=0$.\footnote{Both assumptions are made to keep the
    exposition simple, they are not cricial for
    applicability of the task loss estimation approach.}  
During the training, the learner
is provided with training pairs $(x_i,y_i)$, where
$y_i=g(x_i)$. We assume that given the ground truth $y_i$,
the loss $L(x, \hat{y})$ can be efficiently for any answer
$\hat{y}$.

The training problem is then defined as follows. Given a family of parametrized mappings
$\left\{h_{\alpha}\right\},  \alpha \in \mathcal{A}$ from 
$\mathcal{X}$ to $\mathcal{Y}$, try to choose
one that minimizes (as much as possible) the \textit{risk functional}:
\begin{align}
R(\alpha) = \int\limits_{x} L(x, h_{\alpha}(x)) P(x) dx,
\end{align}
where $P$ is an unknown data distribution. The choice must
be made using only a sample \mbox{$S =
\left\{x_i\right\}_{i=1}^N$} from the distribution $P$ with
ground truth answers $\left\{y_i\right\} _{i=1}^N$ available
for $x_i \in S$.


Here are two examples of task losses that are pretty much standard
in some key supervised learning problems:
\begin{itemize}
\item the 0-1 loss used in classification problems is
$L(x, y) =  
\begin{cases}
1,& g(x) = y \\
0,& g(x) \neq y
\end{cases}$;
\item the Levenshtein distance used in speech
recognition is
$L(x, y) = \rho_{\text{levenstein}}(g(x), y)$ is 
the minimum number of changes required to transform a
transcript $y$ into the correct transcript $g(x)$.
\end{itemize}

\paragraph{Empirical Risk and Surrogate Losses}
\begin{figure}[h]
\centering
    \begin{subfigure}{.5\textwidth}
        \centering
        \includegraphics[width=\linewidth]{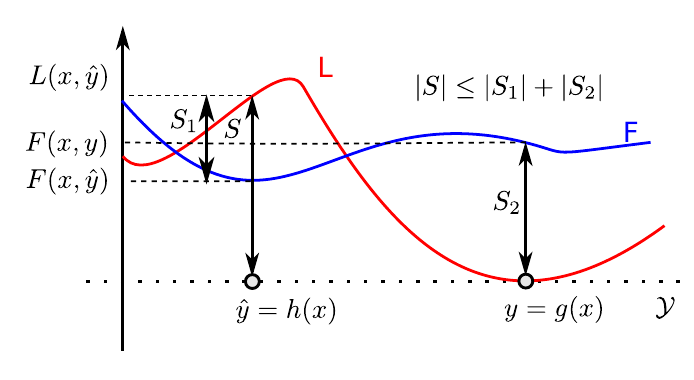}
        \caption{}
        \label{fig:idea1}
    \end{subfigure}%
    \begin{subfigure}{.5\textwidth}
        \centering
        \includegraphics[width=\linewidth]{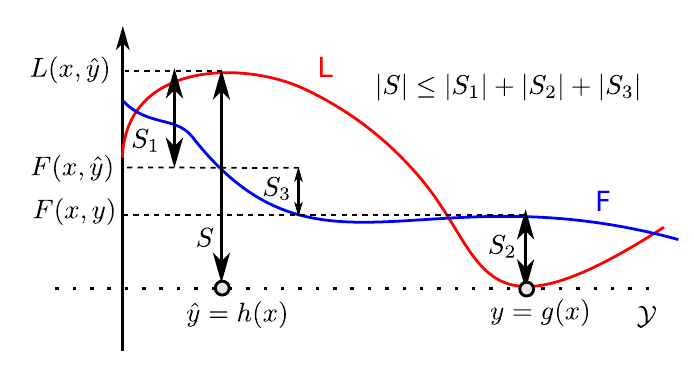}
        \caption{}
        \label{fig:idea2}
    \end{subfigure}
    \caption{A graphical illustration of how the loss
        estimation error 
        provides an upper bound for the task loss $L$, which is
        the underlying idea of Theorem
        \ref{th:bound_for_loss}. The segments $S$, $S_1$,
        $S_2$, $S_3$ on the picture stand for the four main
        terms of the theorem statement, 
        $L(x, \hat{y})$, $|L(x, \hat{y}) - F(x, \hat{y})|$,
        $|F(x, y)|$, $F(x, \hat{y}) - F(x, y)$
        respectively. The location of the segments related
        to each other explains why the loss estimation error
        gives a bound on the task loss $L(x, \hat{y})$ of
        the prediction $\hat{y}$.       
        Figure \ref{fig:idea1} displays the
        case when the mininum of $F(x)$ is successfully
        found by $h_{\alpha}(x)$. Figure \ref{fig:idea2}
        explains the term  $F(x, \hat{y}) - F(x, y)$
        which appears when $h_{\alpha}(x)$ is an approximate
        minimizer incapable to find an output with a score lower
        than $F(x, y)$. }
    \label{fig:idea}
\end{figure}
Under the assumptions that $S$ is big enough 
and the family $\mathcal{A}$ is limited or some form
of regularization is introduced, the
\textit{empirical risk} $\hat{R}(\alpha)$
can be minimized 
\begin{align}
\hat{R}(\alpha) = \frac{1}{N} \sum\limits_{i=1}^N
L(x_i, h_{\alpha}(x_i)),
\end{align} 
instead of $R$ \citep{vapnik1998statistical}.

A common practical issue with minimizing the empirical risk functional
$\hat{R}(\alpha)$ is that $L(x, y)$ is often not differentiable
with respect to $y$, which in turn renders $\hat{R}(\alpha)$
non-differentiable with respect to $\alpha$ and therefore
difficult to optimize. The prevalent
workaround is to define $h_{\alpha}(x)$ as the minimum of a
\textit{scoring function} $F_{\alpha}(x, y)$ (often also
called energy):
\begin{equation*}
h^{\text{min}}_{\alpha}(x) = \argmin_{y} F_{\alpha}(x, y).
\end{equation*}
Parameters $\alpha$ of the scoring function are chosen
to minimize a (technically empirical) \textit{surrogate risk}
$\mathcal{R}(\alpha)$ defined as the average
\textit{surrogate loss}
$\mathcal{L}(x_i, \cdot)$:
\begin{align}
\mathcal{R}(\alpha) = \frac{1}{N} \sum\limits_{i=1}^N
\mathcal{L}(x_i, F_{\alpha}(x_i)),
\label{eq:surrogate}
\end{align}
where $F_\alpha(x_i)\in\mathbb{R}^{|\mathcal{Y}|}$ is the vector of
scores computed on all elements of $\mathcal{Y}$\footnote{Without loss of
generality, we assume here that the output space is discrete.}.

We argue that, for the transition from the empirical risk
$\hat{R}$ to the surrogate risk $\mathcal{R}$ to be
helpful, a number of conditions should hold:
\begin{enumerate}
\item
It must be easy to compute predictions $h_\alpha^{\text{min}}(x)$. Thus
$F_\alpha(x,y)$ must be easy to minimize over $y$, at least in an
approximate sense. For
instance, in most classification problems this is not an
issue at all because the output space $\mathcal{Y}$ is
small. On the other hand, for structured output
prediction this might be a significant issue.
\item 
$\mathcal{R}$ should be simpler to optimize than
$\hat{R}$. 
\item    
Optimization of $\mathcal{R}$ should result in
optimization of $\hat{R}$. 
\end{enumerate}

Let us consider two examples of surrogate losses
\begin{itemize}
\item
The \textit{cross-entropy surrogate loss}
$\mathcal{L}_{CE}$ is applicable when the
scores $F_\alpha(x,y)$ are interpreted as unnormalized negative
log-probabilities:
\begin{align}
    \label{eq:ce}    
    \mathcal{L}_{\text{CE}}(x, F_\alpha(x)) &= 
        F_\alpha(x, g(x)) - 
        \log( \sum\limits_{y' \in \mathcal{Y}}
        \exp(F_\alpha(x, y'))) ,
    \\
    \mathcal{R}_{\text{CE}}(\alpha) &= \frac{1}{N} \sum\limits_{i=1}^N
    \mathcal{L}_{\text{CE}}(x_i, F_{\alpha}(x_i)).
\end{align}
With $\mathcal{L}_{\text{CE}}$ choosing $\alpha$ that minimizes 
$\mathcal{R}_{\text{CE}}(\alpha)$ corresponds to Maximum
Likelihood Estimation (MLE).
\item 
A generalized hinge loss used in Structured Support
Vector Machines \citep{tsochantaridis2005large}:
\begin{align*}
    \mathcal{L}_{\text{hinge}}(x, F_\alpha(x)) = 
    \max\limits_{y} \left(F_\alpha(x, g(x)) -
    F_\alpha(x, y) + L(g(x), y) , 0\right).
\end{align*}
The respective surrogate risk $\mathcal{R}_{\text{hinge}}$ is
defined similarly to $\mathcal{R}_{\text{CE}}$.
\end{itemize}
Both of these surrogate loss functions have properties that make them relatively simple to
optimize. The cross-entropy is both differentiable and convex. The hinge loss is
piecewise differentiable and convex as well.
We refer the reader to \citet{lecun2006tutorial} for a
survey of surrogate loss functions (note that their definition of a loss function differs
slightly from the one we use in this text).

Popular surrogate losses are often agnostic to
the task loss $L$, the cross-entropy
surrogate loss $\mathcal{L}_{\text{CE}}$ being a good example.
Even if we find parameters $\alpha_{\text{CE}}$ which make
the cross-entropy
$\mathcal{R}_{\text{CE}}(\alpha_{\text{CE}})$ arbitrary small, there is no guarantee that the empirical
risk $\hat{R}(\alpha_{\text{CE}})$ will also be small. However, some
surrogate losses, such as the generalized hinge loss
$\mathcal{L}_{\text{hinge}}$, provide certain guarantees for the
empirical risk. Specifically, one can see that $L(x,
h^{\text{min}}_{\alpha}(x))  \leq
\mathcal{L}_{\text{hinge}}(x, F(x))$, which implies
$\hat{R}(\alpha) \leq \mathcal{R}_{\text{hinge}}(\alpha)$, or simply put,
minimizing $\mathcal{R}_{\text{hinge}}$ \textit{necessarily}
pushes $\hat{R}$ down.

\paragraph{Task Loss Estimation}
In this paper we introduce a novel paradigm for building
surrogate losses with guarantees similar to those of
$\mathcal{L}_{\text{hinge}}$. Namely, we propose to interpret
the scoring function $F$ as an estimate of the task loss $L$ itself.
In other words we want $F_{\alpha}(x, y) \approx L(x, y)$.

We can motivate this approach by showing that for the empirical risk to be low, 
it is sufficient for the task loss and the score to be similar at only two
points: 
the ground truth $g(x)$ and the prediction $h_{\alpha}(x)$.
We combine the estimation errors for these two outputs to 
obtain a new surrogate loss
$\mathcal{L}_{\text{min},\text{min}}$ which we
call the $\textit{min-min loss}$.
\begin{theorem} 
    Let $\mathcal{L}_{\text{min}, \text{min}}$ be defined as follows:
\begin{align}
    \mathcal{L}_{\text{min, min}}(L(x), F_{\alpha}(x)) = 
    |F_{\alpha}(x, y)| + |L(x, \hat{y}) - F_{\alpha}(x,
    \hat{y})|,
\end{align}
here $y = g(x)$, $\hat{y} = h_{\alpha}(x)$.
Then the respective surrogate risk $\mathcal{R}_{min,min}$ 
provides the following bound on $\hat{R}$
\begin{align}
    \hat{R}(\alpha) \leq \mathcal{R}_{\text{min, min}}(\alpha)
    + M(\alpha),
\end{align}
where
\begin{align*}
    M(\alpha) = 
    \frac{1}{N} \sum\limits_{i=1}^N 
    \max\left(F(x_i, \hat{y}_i) - F(x_i, y_i), 0\right).
\end{align*}
\label{th:bound_for_loss}
\end{theorem}
Figure \ref{fig:idea} illustrates the statement of Theorem
\ref{fig:idea}. Simply put, the theorem says that if
$h_{\alpha}=h^{min}_{\alpha}$, or if $h_{\alpha}$ is a good
enough approximation of $h^{min}_{\alpha}$ such  that the term
$M(\alpha)$  is small, the surrogate loss
$\mathcal{R}_{\text{min,min}}$ is a sensible substitute for $\hat{R}$.
Please see Appendix for a formal proof of the theorem.

The key difference of our new approach from the generalized
hinge loss is that it assigns a fixed target $L(x, y)$ for the score $F(x,
y)$ of every pair $(x, y) \in \mathcal{X} \times
\mathcal{Y}$. This target is independent of the values 
of $F(x, y')$ for all other $y' \in \mathcal{Y}$. The
knowledge that $L$ is the target can be used at the stage of
designing the model $F_{\alpha}(x, y)$. For example, when
$y$ has a structure, a $L(x, y)$ might be decomposed into
separate targets for every element of $y$, thereby making
optimization of $\mathcal{R}$ more tractable.

In consideration of optimization difficulties, our new
surrogate loss $\mathcal{L}_{\text{min, min}}$ is piece-wise smooth
like $\mathcal{L}_{hinge}$, but it is not convex and even
not continuous. In practice, we tackle the optimization by
fixing the outputs $h_{\alpha}(x)$ for a subset of the
sample $S$, improving $\mathcal{L}_{\text{min, min}}$ with the
fixed outputs by e.g.\ a gradient descent step, and doing the
same iteratively.

\section{Task Loss Estimation for Sequence Prediction}\label{sec:lossestim}

In sequence prediction problems the outputs are 
sequences over an alphabet $C$.
We assume that the alphabet is not too big, more specifically,
that a loop over its elements is feasible. In addition
we extend the alphabet $C$ with a special end-of-sequence
token $\$$, creating the extended alphabet $\overline{C} = C
\cup \{ \$ \}$. For convenience, we assume that all valid
output sequences must end with this token. Now we can
formally define the output space as the set of all sequences which end with 
the end-of-sequence token
$\mathcal{Y} = \{y\$ : y  \in C^*\}$, where $C^*$ denotes a set of all 
finite sequences over the alphabet $C$.

We will now describe how task loss estimation can be applied to sequence
prediction for the following specific scenario:
\begin{itemize}
    \item The score function is an Encoder-Decoder model.
    \item The prediction $h^{\text{min}}_{\alpha}$ is approximated with a beam search or a greedy search.
\end{itemize}

\subsection{Encoder-Decoder Model}

A popular model for sequence prediction is the
Encoder-Decoder model. In this approach, the Decoder is
trained to model the probability 
$P(y^j|z(x), y^{1 \ldots j - 1})$
of the next token $y^j$ given a
representation of the input $z(x)$  produced by the Encoder, and
the previous tokens $y^{1 \ldots j - 1}$,
where $y=g(x)$ is the ground truth output.
Decoders are typically implemented using recurrent neural networks.
Using the terminology of this paper, one can
say that a standard Encoder-Decoder implements a
parametrized function
$\delta_{\alpha}(c, x, y^{1 \ldots j - 1})$ that defines the
scoring function as follows:
\begin{align}
F^{ED1}_{\alpha}(x, y) = 
\sum\limits_{j=1}^{|y|}
- \log q_{\alpha}(y^j, x, y^{1 \ldots j}),
\label{eq:encdec_scoring} \\
q_{\alpha}(y^j, x, y^{1 \ldots j}) =  
\frac
{
    \exp \left( \delta_{\alpha}(y^j, x, y^{1 \ldots j})
    \right)
}
{
    \sum\limits_{c \in \overline{C}} 
    \exp \left( \delta_{\alpha}(c, x, y^{1 \ldots j}) \right)
}.
\label{eq:normalization}
\end{align}
The cross-entropy surrogate loss can be used for training
Encoder-Decoders. Since the score function
\eqref{eq:encdec_scoring} defined by
an Encoder-Decoder is a proper distribution, the exact
formula for the surrogate loss is simpler than in Equation
\ref{eq:ce}
\begin{align*}
\mathcal{L}_{CE}(x, F^{ED1}_{\alpha}(x)) 
= F^{ED1}_{\alpha}(x, y) = 
\sum\limits_{j=1}^{|y|}
- \log q_{\alpha}(y^j, x, y^{1 \ldots j - 1}), \\
\textrm{ where } y = g(x).
\end{align*}

Exactly computing $h^{\text{min}}_{\alpha}$ is not possible for
Encoder-Decoder models. A beam search procedure
is used to compute an approximation $h^{B}_{\alpha}$, where
$B$ is the beam size. In beam search at every step $k$ the
beam, that is a set of
$B$ ``good prefixes'' of length $k$, is replaced by a set of
good prefixes of length $k+1$. The transition is done by
considering all continuations of the beam's sequences
and leaving only those $B$ candidates for which the partial sum of
$\log q_{\alpha}$ is minimal.

\begin{figure}
\centering
\includegraphics[width=0.5\textwidth]{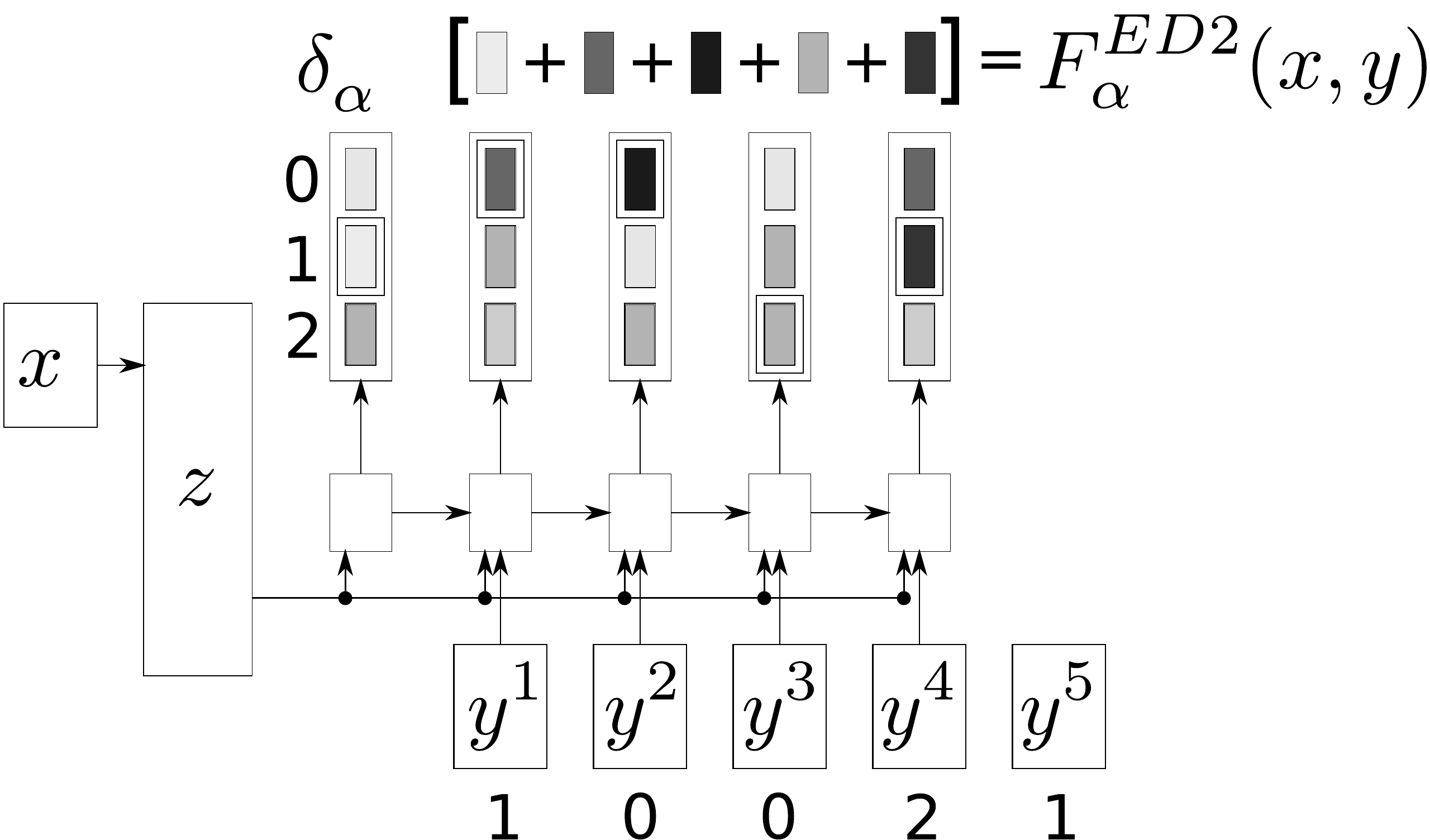}
\caption{A schematic representation of an Encoder-Decoder
    architecture implementing the score function $F^{ED2}_{\alpha}(\cdot)$. 
    For this example, the score
of a sequence of labels $\{y^1,\cdots,y^5\}$ and an input sequence $x$ is computed, where each label $y^j$ is from
the alphabet $\{0,1,2\}$. For each label in the sequence, the decoder produces a
vector $\delta_{\alpha}$ that represents the predicted
change $\delta_o$ in the optimistic loss for each
possible symbol at the next time step. The score for the
whole sequence is
computed by summing  $\delta_{\alpha}(y^j,y^{1\cdots
j-1}, x)$ for all $j$. Note that at each timestep, the
decoder also uses
the representation $z(x)$ computed by the encoder.}
\label{fig:encdec}
\end{figure}

\subsection{Applying Task Loss Estimation to Encoder-Decoders}

\paragraph{Adapting the Min-Min Loss.}
We want to keep the structure of the scoring function
defined in Equation \eqref{eq:encdec_scoring}. However, the
normalization carried out in \eqref{eq:normalization} is not
necessary any more, so our new scoring function is simply the
sum of $\delta_{\alpha}$:
\begin{align*}
F^{ED2}_{\alpha}(x, y) = 
\sum\limits_{j=1}^{|y|}
\delta_{\alpha}(y^j, x, y^{1 \ldots j - 1}).
\end{align*}
Now, in theory, the min-min loss $\mathcal{L}_{\text{min, min}}$
could be used for training $F^{ED2}_{\alpha}$. However,
there are two concerns which render this straight-forward
approach less attractive:
\begin{itemize}
\item
    Intuitively, constraining only the sum of
    $\delta_{\alpha}$ might provide not enough
    supervision for training. Namely, the gradient of
    $\mathcal{L}_{\text{min, min}}$ would be the same with
    respect to all 
    $\delta_{\alpha}(y^j, x, y^{1 \ldots j - 1})$, which
    might hamper gradient-based optimization methods.
\item       
    There is no guarantee that the beam search will be
    able to work with $\delta_{\alpha}$ values learnt
    this way.
\end{itemize}
To circumvent both of these potential issues, we propose to
break the target loss  $L(x, y)$  into subtargets
$\delta_L^j(x, y)$ assigned
token-wise. We define
the \emph{optimistic task loss} $L_o(x, y)$ for an output
prefix $y$ as the loss of the best
possible continuation of the prefix $y$. For completed
output sequences, that is those ending with the
end-of-sequence token, we say that the optimistic task loss is equal
to the task loss. This results in the following formal
definition:
\begin{align}
    L_o(x, y) = 
    \begin{cases}
        \min\limits_{z \in B^*} L(x, yz\$), & y \in C^*; \\
        L(x, y), & y \in \mathcal{Y},
    \end{cases} 
\end{align}
We argue that the change of the optimistic task loss 
$\delta_o(y^j, x, y^{1 \ldots j - 1}) = L_o(x, yc) - L_o(x, y)$ is a
good target for $\delta_{\alpha}(y^j, x, y^{1 \ldots j - 1})$.
Indeed,  the pruning during beam search is done by looking at the sum
$s(x, y^{1 \ldots k}) = \sum\limits_{j=1}^k
\delta_{\alpha}(y^j, x, y^{j-1})$ 
for the prefixes $y$ from the beam. Informally, the
pruning procedure should remove prefixes whose continuations are
unlikely to be beneficial. The optimistic loss
$L_o(x, y)$ tells us what is the lowest loss one can obtain by
continuing $y$ in an arbitrary way, and hence, it can be used for
selecting the prefixes to be continued. Assuming that the
network learns to output 
$\delta_{\alpha}(c, x, y^{1 \ldots j}) \approx 
\delta_o(c, x, y^{1 \ldots j})$, we can hope that
pruning  by
$s_k(x, y^{1 \dots j}) \approx L_{\text{opt}}(x, y^{1 \dots k})$ will keep the good prefixes in.

Our new surrogate loss consisting of the sum of token-wise
errors looks as follows: 
\begin{align}
\mathcal{L}^{ED}_{\text{min, min}}(x, \delta_{\alpha}(x)) =
\sum\limits_{j=1}^{|y|} 
|\delta_{\alpha}(y^j, x, y^{1 \ldots j - 1}) - 
 \delta_o(y^j, x, y^{1 \ldots j - 1})|  \\ + 
 \sum\limits_{j=1}^{|\hat{y}|} 
 |\delta_{\alpha}(\hat{y}^j, x, \hat{y}^{1 \ldots j - 1}) - 
 \delta_o(\hat{y}^j, x, \hat{y}^{1 \ldots j - 1})| 
,
\end{align}
where $y=g(x)$, $\hat{y} = h^{\text{min}}_{\alpha}(x)$. Note, that
$\mathcal{L}^{ED}_{\text{min,min}}$ extends our previous surrogate
loss definition from \eqref{eq:surrogate} by working not on
$F_{\alpha}(x)$ but on its
additive components $\delta_{\alpha}(y^j, x, y^{1 \ldots j -
1})$. One can also see that $\mathcal{L}^{ED}_{\text{min,
min}}(x, \delta_{\alpha}(x))
\geq \mathcal{L}_{\text{min, min}}(x, \delta_{\alpha}(x))$ because of 
the triangle inequality, which implies that the respective
surrogate risk is a bound for the empirical risk $\mathcal{R}^{ED}_{\text{min, min}} \geq
\hat{R}(\alpha)$.

A careful reader might have noticed, that in practice we do not
have access to $\mathcal{L}^{ED}_{\text{min, min}}$, because we can
not compute $h^{\text{min}}_{\alpha}(x)$. The best we can have is
$\mathcal{L}^{ED}_{\text{min}, B}(x, y)$ defined in a similar way
but using the beam search to compute $\hat{y} =
h^B_{\alpha}(x)$ instead of the intractable exact
minimization. However, according to Theorem \ref{th:bound_for_loss} 
minimizing $\mathcal{L}^{ED}_{\text{min}, B}$ guarantees low
empirical risk for beam search predictions $h^B_{\alpha}(x)$, as long as the
beam search finds an output with a score that is lower than the score of the
ground truth. In our experience, this is usually the case for
Encoder-Decoder models.

\paragraph{A Loss for the Greedy Search}
One disadvantage of the approach with
$\mathcal{L}^{ED}_{\text{min}, B}$ is that computing the surrogate
loss, and therefore also its gradients, becomes
quite expensive. We address this issue by proposing another
surrogate loss which only involves beam search with the beam size $B=1$, also
often called greedy search. The new surrogate loss
$\mathcal{L}^{ED}_{\text{greedy}}$ is defined as follows:
\begin{align}
    \mathcal{L}^{ED}_{\text{greedy}}(x, \delta_{\alpha}(x)) = 
    \sum\limits_{j=1}^{|\hat{y}|} 
    |\delta_{\alpha}(\hat{y}^j, x, \hat{y}^{1 \ldots j - 1}) - 
    \delta_o(\hat{y}^j, x, \hat{y}^{1 \ldots j - 1})|
    + 
    |\delta_{\alpha}(c^j_{\text{min}}, x, \hat{y}^{1 \ldots
        j - 1})| ,
    \label{eq:greedy_loss}
\end{align}
where $\hat{y} = h^1_{\alpha}(x)$, $c^j_{\text{min}} =
\argmin\limits_{c \in \overline{C}} \delta_{o}(c, x, y^{1 \ldots j
    - 1})$.  We can show, that
optimizing the respective surrogate risk 
$\mathcal{R}^{ED}_{\text{greedy}}$ necessarily improves
the performance of greedy search:
\begin{theorem}
    The empirical risk $\hat{R}_{\text{greedy}}$ associated
    with using $h^1_{\alpha}$ for giving predictions is
    bounded by $\mathcal{R}^{ED}_{\text{greedy}}$, that is
    $\hat{R}_{\text{greedy}}(\alpha) \leq
    \mathcal{R}^{ED}_{\text{greedy}}(\alpha)$.
    \label{th:greedy}
\end{theorem}
The proof can be found in the Appendix. Now, with the greedy search,
the gradient of $\hat{R}_{\text{greedy}}(\alpha)$ can be computed
just as fast as the gradient of the average cross-entropy, since
the computation of the gradient can be combined with the
search. 

\paragraph{Tricks of the Trade}

Driven by our intuition about the training process we make
two further changes to the loss
$\mathcal{L}_{\text{greedy}}$. First, we change Equation \ref{eq:greedy_loss}
by adding all characters into consideration:
\begin{align}
    \mathcal{L}^{ED}_{\text{greedy1}}(x, \delta_{\alpha}(x)) =
    \sum\limits_{j=1}^{|\hat{y}|} 
    \sum\limits_{c \in \overline{C}}
    |\delta_{\alpha}(c, x, \hat{y}^{1 \ldots j - 1}) - 
    \delta_o(c, x, \hat{y}^{1 \ldots j - 1})| 
.
    \label{eq:greedy_loss1}
\end{align}
Our reasoning is that by providing a more informative training
signal at each step we help optimization. We note, that the
bound on the empirical risk provided by the respective
surrogate risk $\mathcal{R}^{ED}_{\text{greedy1}}(\alpha)$ is
looser then the one by
$\mathcal{R}^{ED}_{\text{greedy}}(\alpha)$ since 
$\mathcal{R}^{ED}_{\text{greedy}} \leq \mathcal{R}^{ED}_{greedy1}$.
On the other hand, $\mathcal{R}^{ED}_{\text{greedy1}}$ enforces
a margin between the best next token and all the worse ones,
which can possibly help generalization.

Finally, we found $\mathcal{L}^{ED}_{\text{greedy1}}$ hard to
optimize because the gradient of $|a - b|$ is always either
+1 or -1, that is it does not get smaller when $a$ and $b$
approach each other. To tackle this we replaced the absolute
value by the square:
\begin{align}
    \notag
    \mathcal{L}^{ED}_{\text{greedy2}}(x, \delta_{\alpha}(x)) =
    \sum\limits_{j=1}^{|\hat{y}|} 
    \sum\limits_{c \in \overline{C}}
    (\delta_{\alpha}(c, x, \hat{y}^{1 \ldots j - 1}) - 
    \delta_o(c, x, \hat{y}^{1 \ldots j - 1}))^2
.
\end{align}

\paragraph{Example: Edit Distance}

We explain how the decomposition of the task loss $L(x, y)$
into a sum $\sum\limits_{j=1}^{|y|} \delta_o(y^j, x, y^{1
    \ldots j - 1})$ works on the example of the edit distance.
The edit distance $\rho_{\text{levenstein}}(s_1, s_2)$ between two strings $s_1,
s_2 \in C^*$ is the minimal number of actions required to
transform $s_1$ into $s_2$, where the actions allowed are
token deletion, insertion and substitution. If the loss
$L(x, y)$ is defined as the edit distance 
$\rho_{\text{levenstein}}(g(x), y)$, there is a compact expression for the
optimistic loss $L_o(x, y)$:
\begin{align}
    L_o(x, y) = 
    \begin{cases}
        \min_{k=0}^{k=|g(x)|} \rho_{\text{levenstein}}(y, g(x)^{1 \ldots k}), & y
        \in C^*, \\
        \rho_{\text{levenstein}}(y, g(x)), & y \in \mathcal{Y}.
    \end{cases}
    \label{eq:opt_loss_ed}
\end{align}
Equation \eqref{eq:opt_loss_ed} formalizes the consideration
that the optimal way to continue a prefix $y$ is to append a
suffix of the ground truth $g(x)$. From the obtained
expression for $L_o(x, y)$ one can see that $\delta_o(c, x,
y)$ can only be 0 or -1 when $c \neq \$$. Indeed, by
definition $\delta_o \geq 0$, and also adding a
character $c$ to a prefix $y$ can only change the edit
distance $\rho(y, g(x)^{1 \ldots k})$ by 1 at most. For the case of
$c=\$$ the value $\delta_o(\$, x, y)$ can be an arbitrarily
large negative number,
in particular for prefixes $y$ which are shorter then
$g(x)$. It would be a waste of the model capacity to try to
exactly approximate such larger numbers, and in practice we
clip the values $\delta_o(\$, x, y)$ to be at most -5.

An attentive reader might have noticed, that for complex
loss functions such as e.g. BLEU and METEOR computing
the loss decomposition like we did it above might be
significantly harder. However, we believe that by
considering all
ground truth suffixes one can often find a close to optimal
continuation.  

\section{Related Work}\label{sec:related}

In an early attempt to minimize the empirical risk for speech recognition models, word
error rate scores were used to rescale a loss similar to the objective that is
often referred to as Maximum Mutual Information \citep{povey2002}.
For each sequence in the data, this objective requires a summation over all
possible sequences to compute the expected word error rate from the groundtruth, something that is
possible for only a restricted class of models. A recent survey
\citep{he2008discriminative} explains and documents improvements in speech
recognition brought by other methods of discriminative training of
speech recognition systems.

In the context of Encoder-Decoders for sequence generation, a curriculum
learning \citep{bengio2009curriculum} strategy has been proposed to address the
discrepancy between the training and testing conditions of models trained with
maximum likelihood \citep{bengio2015scheduled}. It was shown that the
performance on several sequence prediction tasks can be improved by gradually transitioning
from a fully guided training scheme to one where the model is increasingly
conditioned on symbols it generated itself to make training more similar to the
decoding stage in which the model will be conditioned on its own predictions as well.
While this approach has an intuitive appeal and clearly works well in some
situations, it doesn't take the task loss into account and to our knowledge no
clear theoretical motivation for this method has been provided yet.
Another issue is that one needs to decide how fast to transition between the two
different types of training schemes.

Recently, methods for direct empirical risk minimization for structured
prediction have been proposed that treat the model and the approximate
inference procedure as a single black-box method for generating predictions
\citep{stoyanov2011,domke2012}. The gradient of the loss is backpropagated
through the approximate inference procedure itself.  While this approach is
certainly more direct than the optimization of some auxiliary loss, it requires
the loss to be differentiable.

\citet{hazan2010} propose a method for direct loss minimization that
approximates the gradient of the task loss using a loss adjusted inference
procedure. This method has been extended to Hidden Markov Models and applied to
phoneme recognition \citep{keshet2011direct}.

For a model that provides a distribution over structured output configurations,
the gradient with respect to any expectation over that distribution can be
estimated using a sampling approach. This technique has been used for
speech recognition \citep{graves2014towards} to estimate the
gradient of the transcription loss (i.e., the word error rate) and is
equivalent to the REINFORCE method \citep{williams1992simple} from reinforcement learning.
A downside of this method is that in many cases the gradient estimates
have high variance. The method also assumes that it is possible and
computationally feasible to sample from the model.
A related approach is to use an inference method to generate a list of the $n$
best candidate output predictions according to the model (note that for this the
model doesn't need to be probabilistic) and approximate the expected loss using
an average over these candidate predictions \cite{gao2013training}. Similarly,
one can anneal from a smooth expectation approximated with a large number of
candidates towards the loss of a single prediction \cite{smith2006minimum}.

\section{Experimental Setup and Results}
\label{seq:results}
For experimental confirmation\footnote{Our code is available
    at https://github.com/rizar/attention-lvcsr} of the theory discussed in 
Sections~\ref{sec:lossdefs} and \ref{sec:lossestim}, we use
a character-level speech 
recognition task similar to~\cite{bahdanau2015end-to-end}.
Like in our previous work, we used the Wall Street Journal (WSJ) speech corpus for our experiments. The model is trained on the full 81 hour 'train-si284' training set, we use the 'dev93' development set for validation and model selection, and we report the performance on the 'eval92' test set. The inputs to our models were sequences of feature vectors. Each feature vector contained the energy and 40 mel-filter bank features
with their deltas and delta-deltas, which means that the
dimensionality of the feature vector is 123. We use the
standard trigram language model shipped with the WSJ
dataset; in addition we experiment with its extended version
created by Kaldi WSJ s5 recipe \citep{povey2011kaldi}.

Our main baseline is an Encoder-Decoder from our previous
work on end-to-end speech recognition
~\citep{bahdanau2015end-to-end}, trained with the
cross-entropy surrogate loss. We trained a model with the
same architecture but using the task loss estimation
$\mathcal{L}^{ED}_{\text{greedy2}}$ criterion, which
involves greedy prediction of the candidate sequence $\hat{y}$ during
training.  Algorithm \ref{alg} formally describes our
training procedure. 

Our main result is the 13\% relative improvement of
Character Error Rate that 
task loss estimation training brings compared to the
baseline model when no external language model is used
(see Table~\ref{tab:results}). This setup, being not typical
for speech recognition research, is still an interesting
benchmark for sequence prediction algorithms. We note, that 
the Word Error Rate of 18\% we report here is the best in
the literature. Another class of models for which results
without the language model are sometimes reported are 
Connectionist Temporal Classification (CTC) models 
~\citep{graves2014towards, miao2015eesen, hannun2014first},
and the best result we are aware of is 26.9\% reported by
\cite{miao2015eesen}.

In our experiments with the language models we linearly
interpolated the scores produced by the neural networks with
the weights of the Finite State Transducer (FST), similarly
to \citep{miao2015eesen} and
\citep{bahdanau2015end-to-end}. Addition of language models 
resulted in  a typical large performance improvement, 
but the advantage over the cross-entropy trained model was
largely lost. Both the baseline and the experimental model perform
worse than a combination of a CTC-trained network and a
language model. As discussed in our previous work
\citep{bahdanau2015end-to-end}, we attribute it to 
the overfitting from which Encoder-Decoder models suffers
due to their implicit language modelling capabilities.

\begin{algorithm}
    \While{$L^{ED}_{greedy2}$ improves on the validation set}{
        fetch a batch of input sequences $B$\;
        generate $\hat{y}_i$ for each $x_i$ from $B$ using
        the greedy search\;
        compute the score components $\delta_{\alpha}(c, x_i, \hat{y}_i^{1 \ldots
            j-1})$ \;
        compute the component-wise targets 
        $\delta_o(c, x_i, \hat{y}_i^{1 \ldots j - 1})$ as
        changes of the optimistic task loss\;
        $L^{ED}_{greedy2} = \frac{1}{|B|} 
            \sum\limits_{i=1}^{|B|}
            \sum\limits_{j=1}^{|\hat{y}|} 
            \sum\limits_{c \in \overline{C}}
            \left(\delta_{\alpha}(c, x_i, \hat{y}_i^{1 \ldots j - 1}) - 
            \max\left( \delta_o(c, x_i, \hat{y}_i^{1 \ldots j -
                    1}), -5 \right)\right)^2$\;
        compute the gradient of
        $L^{ED}_{greedy2}$ and update the  parameters $\alpha$;
    }
    \caption{The training procedure used in our experiments.
        Note, that generation of $\hat{y}_i$ and gradient
        computation can be combined in an efficient
        implementation, making it exactly as fast as
        cross-entropy training.}
    \label{alg}
\end{algorithm}

It is notable, that the performance of the experimental
model changes very little when we change the beam size from
10 to 1.  An unexpected result of our experiments is that
the sentence error rate for the loss estimation model is
consistently lower.  Cross-entropy is de-facto the standard
surrogate loss for classifiers, and the sentence error rate
is essentially the classification error, for which reasons
we did not expect an improvement of this performance
measure. This result suggests that for classification
problems with very big number of classes the cross-entropy might
be a non-optimal surrogate loss.

\begin{table}
    \centering
    \caption{Character, word, and sentence error rates (CER, WER, and SER) for 
        the cross-entropy (CE) and the task loss estimation (TLE) models. 
        The first three sections of the table present
        performance of the considered models with no language model integration, 
        with a standard trigram language model (std LM), and
        with an extended language model (ext LM). The last
        section contains results from
        \cite{graves2014towards} and \cite{miao2015eesen}.
		We found that increasing the beam size over 100 for
        the CE model does not 
        give any improvement. In addition to the results on the test set (eval92) we 
        reported the performance on the validation set (dev93).
    }
    \begin{tabular}{l l | c c c || c c c }
        Model & \pbox{20cm}{Beam \\ size} & \multicolumn{3}{c||}{Eval92 set} & \multicolumn{3}{c}{Dev93 set} \\ 
          &                          &  CER\% & WER\% &
          SER\% &  CER\% & WER\% & SER\%\\
    \hline
    \hline
    CE, no LM &  1   & 7.6 & 21.3 & 89.4 & 8.8 & 23.8 & 90.3\\
    TLE, no LM  &  1 & 6.1 & 18.8 & 86.6  & 7.8 & 23.0 & 91.9 \\
    CE, no LM  &  10   & 6.8 & 19.5 & 87.8  & 9.0 & 23.9 & 90.3\\
    TLE, no LM &  10 & 5.9 & 18.0 & 86.2  & 7.6 & 22.1 & 89.9 \\
	\hline
	CE + std LM & 100 & 4.8 & 10.8 & 63.4 & 6.5 & 14.6 & 75.0\\
	TLE + std LM & 100 & 4.4 & 10.5 & 64.6 & 6.0 & 14.2 & 73.8\\
	\hline
	CE + ext LM & 100 & 3.9 & 9.3 & 61.0 & 5.8 & 13.8 & 73.6\\
	TLE + ext LM & 100 & 4.1 & 9.6 & 62.2 & 5.7 & 13.7 & 74.2\\
	TLE + ext LM & 1000 & 4.0 & 9.1 & 61.9 & 5.4 & 12.9 & 73.2\\
	\hline
	\hline
	Graves et al., CTC, no LM & $-$ & $-$ & 27.3 & $-$  & $-$ & $-$ & $-$\\
	Miao  et al., CTC, no LM & $-$ & $-$ & 26.9 & $-$  & $-$ & $-$ & $-$\\
	Miao et al., CTC + LM & $-$ & $-$ & 9.0 & $-$  & $-$ & $-$ & $-$\\
	Miao et al., CTC + ext LM & $-$ & $-$ & 7.3 & $-$ & $-$ & $-$ & $-$
    \end{tabular}
    \label{tab:results}
\end{table}

\section{Conclusion and Discussion}

The main contributions of this work are twofold. First, we have developed a
method for constructing surrogate loss functions that provide guarantees about
the task loss. Second, we have demonstrated that such a surrogate loss for
sequence prediction performs better than the cross-entropy surrogate loss at
minimizing the character error rate for a speech recognition task.

Our loss function is somewhat similar to the one used in the Structured
SVM \citep{tsochantaridis2005large}. The main difference is
that while the structured SVM uses the task loss to define
the \emph{difference} between the energies assigned to the
correct and incorrect predictions, we use the task loss to
directly define the desired score for all outputs.
Therefore, the target value for the score of an output does
not change during training. 

We can also analyze our proposed loss from the perspective of score-landscape shaping
\citep{lecun2006tutorial}. Maximum likelihood loss applied to
sequence prediction pulls down the score of correct sequences,
while directly pulling up on the score of sequences differing in only
one element. This is also known as teacher-forcing -- the model is
only trained to predict the next element of a correct prefixes of
training sequences. In contrast, our proposed loss function defines
the desired score level for all possible output sequences. Thus it
is not only possible to train the model by lowering the score of the
correct outputs and raising the score of neighboring incorrect
ones, but by precisely raising the score of any incorrect
one. Therefore, the model can be trained on its own mistakes.

Future work should investigate the applicability of our
framework to other task loss functions like the BLEU score.
Our results with the language models stress the importance
of developing methods of joint training of the whole system,
including the language model.  Finally, theoretical work
needs to be done to extend our framework to different
approximate inference algorithms as well and to be able to
make stronger claims about the suitability of the surrogate
losses for gradient-based optimization.

{\bf Acknowledgments}: We thank the developers of
Theano~\citep{Bastien-Theano-2012} and Blocks~\citep{blocksfuel} for
their great work. We thank NSERC, Compute Canada, Canada Research
Chairs, CIFAR, Samsung, Yandex, and National Science Center (Poland) for their support.
We also thank Faruk Ahmed and David Krueger for valuable
feedback.

\bibliography{iclr2016_conference}
\bibliographystyle{iclr2016_conference}

\newpage
\section*{Appendix}

\paragraph{Proof for Theorem~\ref{th:bound_for_loss}}
\begin{proof}
    As illustrated at Figure~\ref{fig:idea}, 
    \begin{equation*}
        L(x, \hat{y}) \leq 
        \begin{cases}
            F(x, y) + |L(x, \hat{y}) - F(x, \hat{y})|, & \text{if }  F(x, \hat{y}) \leq F(x, y), \\
            F(x, y) + |L(x, \hat{y}) - F(x, \hat{y})| + F(x, \hat{y}) - F(x, y), & \text{otherwise}.
        \end{cases}
    \end{equation*}
    Or simplifying 
    \begin{equation*}
        L(x, \hat{y}) \leq F(x, y) + |L(x, \hat{y}) - F(x, \hat{y})| + \max(F(x, \hat{y}) - F(x, y), 0).
    \end{equation*}

    Finally, after summation over all examples $x_i$
    \begin{align*}
        \hat{R}(\alpha) \leq \mathcal{R}_{\text{min,min}} + M(\alpha).
    \end{align*}

\end{proof}

\paragraph{Proof for Theorem~\ref{th:greedy}}
\begin{proof}
    Let us prove the following inequality
    \begin{align}
        \delta_o^{1\ldots j} \leq 
        |\delta_{\alpha}^{1\ldots j} - \delta_o^{1\ldots j}|
        + 
        |\delta_{\alpha}(c^j_{min}, x, y^{1\ldots j-1} )|,
        \label{eq:loss_diff_bound}
    \end{align}
    where we denote $\delta_o^{1\ldots k}=\delta_o(y^k, x, y^{1\ldots k-1})$.

    Equation~\eqref{eq:loss_diff_bound} immediately follows from 
    Theorem~\ref{th:bound_for_loss} when we apply it to every step of loss estimation.
    Then we compute sum over $j=1\ldots |y|$ in Equation~\eqref{eq:loss_diff_bound}, deltas
    sum to $L_o(x, y) = L(x, y)$, which gives
    \begin{align}
        L(x, y) \leq 
        \sum_{j=1}^{|y|}
        |\delta_{\alpha}^{1\ldots j} - \delta_o^{1\ldots j}|
        + 
        |\delta_{\alpha}(c^j_{min}, x, y^{1\ldots j-1})|,
    \end{align}
    which proves the theorem.
\end{proof}

\end{document}